\documentclass[11pt]{article}

\usepackage[final]{acl}

\usepackage{times}
\usepackage{latexsym}

\usepackage[T1]{fontenc}

\usepackage[utf8]{inputenc}

\usepackage{microtype}

\usepackage{inconsolata}

\usepackage{graphicx}

\usepackage{booktabs}
\usepackage{multirow}
\usepackage{amssymb}
\usepackage{CJKutf8}
\usepackage{enumitem}
\usepackage{algorithm}
\usepackage{algorithmic}
\usepackage{amsmath}
\usepackage{tcolorbox}

\newtcolorbox{promptbox}{
    colback=white,
    colframe=black!70,
    boxrule=0.4pt,
    arc=0pt,
    left=1.2mm,
    right=1.2mm,
    top=1.2mm,
    bottom=1.2mm,
    boxsep=0pt,
    fontupper=\small
}

\title{A Multi-Stage Framework for Kuzushiji Character Recognition in Japanese Historical Documents}

\author{
  \textbf{Rui-Yang Ju\textsuperscript{1}},
  \textbf{Kohei Yamashita\textsuperscript{2}},
  \textbf{Hirotaka Kameko\textsuperscript{2}},
  \textbf{Shinsuke Mori\textsuperscript{2}}
\\
\\
  \textsuperscript{1}Graduate School of Informatics, Kyoto University, Kyoto, Japan
\\
  \textsuperscript{2}Academic Center for Computing and Media Studies, Kyoto University, Kyoto, Japan
\\
  \small{jryjry1094791442@gmail.com; forest@i.kyoto-u.ac.jp}
}

\begin{document}
\begin{CJK}{UTF8}{min} 
\maketitle
\begin{abstract}
Kuzushiji was a widely used cursive writing system in pre-modern Japan.
Due to simplification and glyph variation, most modern Japanese readers cannot read Kuzushiji characters.
Consequently, recent studies have developed optical character recognition (OCR) systems for Kuzushiji.
Despite recent progress, Kuzushiji character recognition (KCR) in Japanese historical documents remains challenging because of seal-character overlap and complex layouts, which interfere with character recognition and hinder accurate reconstruction of the reading order.
To address these challenges, we propose a multi-stage KCR framework comprising character detection, cropping, classification, ordering, and large language model (LLM)-based post-OCR correction.
Specifically, we employ a synthetic data augmentation strategy to improve character detection robustness against seal interference and introduce an adaptive column clustering algorithm to reconstruct the reading order.
Finally, we leverage the contextual capabilities of the LLM to correct OCR errors.
In addition, we correct annotation omissions, reconstruct the benchmark dataset, and introduce a synthetic test set with simulated seal interference and an out-of-domain (OOD) test set for evaluation.
Compared with the conventional character-level OCR baseline, our framework achieves relative CER reductions of 43.48\%, 46.02\%, and 39.11\% on the real, synthetic, and OOD test sets, respectively.
\end{abstract}

\begin{figure*}[t]
\centering
\includegraphics[width=\linewidth]{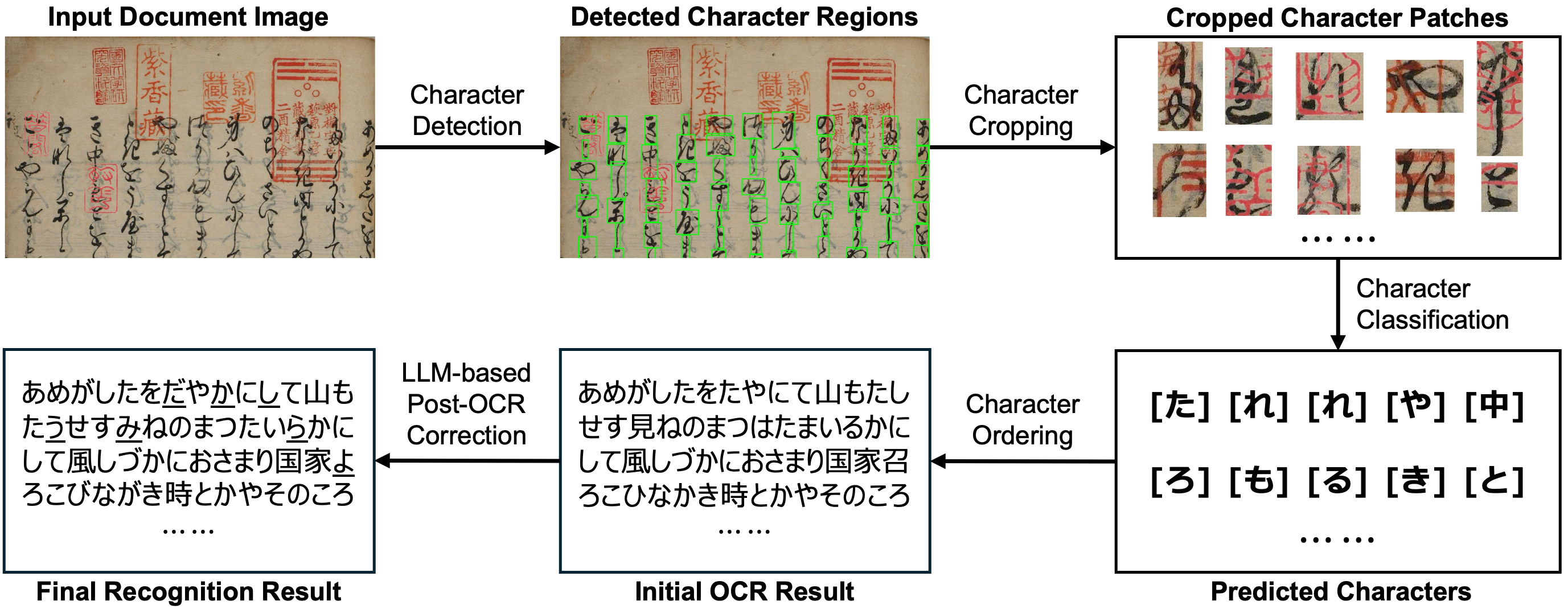}
\caption{Overview of the proposed multi-stage KCR framework for Japanese historical documents, comprising five stages: character detection, cropping, classification, ordering, and LLM-based post-OCR correction.}
\label{fig:pipeline}
\end{figure*}

\section{Introduction}
Kuzushiji (くずし字), a traditional cursive writing system widely used in pre-modern Japan, is prevalent in Japanese historical documents.
Although it shares the same character system as modern Japanese, including Kanji (漢字, Chinese-origin logographic characters) and Kana (仮名, Japanese syllabaries), its character forms differ substantially from their modern counterparts.
Kuzushiji also encompasses multiple cursive writing conventions and numerous variant forms of both Kanji and Kana~\cite{NIJL2018Kuzushiji}.
Following the Meiji Restoration, the adoption of modern printing technologies and educational systems led to a decline in its use~\cite{hashimoto2017kuzushiji}.
As a result, most modern Japanese readers cannot read documents written in Kuzushiji, and only a limited number of trained specialists can accurately interpret them.

Recent advances in deep learning have enabled automated optical character recognition (OCR) of Kuzushiji using models trained on expert-annotated datasets~\cite{clanuwat2018deep}.
To date, several Kuzushiji OCR systems and applications have been developed to assist researchers and the general public in interpreting Japanese historical documents, including Komonjo Camera~\cite{toppan2023fuminoha}, developed by TOPPAN Inc.; KuroNet~\cite{clanuwat2019kuronet,lamb2020kuronet} and miwo~\cite{clanuwat2021miwo}, developed by the Center for Open Data in the Humanities (CODH); and NDLkotenOCR~\cite{kiyonori2023enhancing} and NDLkotenOCR-Lite~\cite{toru2024development}, developed by the National Diet Library (NDL).

Japanese historical documents frequently contain seals that indicate ownership, identity, or provenance.
Because seals may have been added by successive owners and collectors, a single page may contain multiple seals, some of which overlap with handwritten Kuzushiji characters.
Such overlap can interfere with character detection, yet existing Kuzushiji OCR systems do not explicitly address this problem.
Data augmentation~\cite{ghiasi2021simple,ling2022humans} provides a practical approach to improving the robustness of character detectors against such interference.
In addition, Japanese historical documents vary widely in page layout, with some exhibiting complex or irregular layouts.
Although Kuzushiji is conventionally written in vertical columns, with characters read from top to bottom and columns from right to left, many pages contain multiple spatially separated blocks with varying column positions and widths.
These layout variations make it difficult to determine column boundaries and reconstruct the correct reading order.
To address these challenges, we develop a multi-stage Kuzushiji character recognition (KCR) framework. The pipeline is shown in Figure~\ref{fig:pipeline}.

The main contributions of this work are as follows:
(a) We examine 1,000 page images from 13 Japanese historical books, correct annotation omissions in 267 of them, reconstruct the benchmark dataset, and construct both a synthetic test set simulating seal interference and a book-disjoint out-of-domain (OOD) test set for evaluating generalization to unseen books.
(b) We propose a multi-stage KCR framework to improve recognition accuracy and facilitate access to Japanese historical documents for researchers and the general public.
(c) We introduce a synthetic data augmentation (SDA) strategy for character detection, improving robustness to severe seal interference without modifying the detector architecture.
(d) We develop a training-free character ordering algorithm, termed adaptive column clustering (AdaCC), that improves reading-order reconstruction over a supervised LightGBM-based baseline without requiring additional training.
(e) We evaluate LLMs accessed via APIs for post-OCR correction and demonstrate their effectiveness in further reducing recognition errors.


\section{Related Work}\label{sec:related}
Conventional Kuzushiji OCR methods can be broadly categorized into two types: character-level and text-line-level methods.
Character-level methods recognize individual characters and typically predict their locations and corresponding character labels from Japanese historical document images.
The recognized characters are subsequently ordered according to their spatial arrangement to produce the final recognition result.
Representative systems include KuroNet~\cite{clanuwat2019kuronet,lamb2020kuronet} and miwo~\cite{clanuwat2021miwo}, both developed by the CODH.
KuroNet employs a residual U-Net architecture~\cite{ronneberger2015u} to jointly predict the locations and identities of individual characters directly from document images.
Meanwhile, miwo is an application that provides users with access to Kuzushiji OCR models through the CODH server.

The second category comprises text-line-level methods.
These methods first perform layout recognition to detect Kuzushiji text lines in Japanese historical document images.
The detected text lines are then recognized and converted into machine-readable text, after which their reading order is determined to produce the recognition result.
Representative systems include NDLkotenOCR~\cite{kiyonori2023enhancing}, a publicly available OCR system for Japanese historical documents developed by the NDL, and its lightweight variant, NDLkotenOCR-Lite~\cite{toru2024development}, designed for efficient processing in CPU-only environments.
Specifically, NDLkotenOCR employs Cascade Mask R-CNN~\cite{cai2019cascade} for layout recognition and TrOCR~\cite{li2023trocr} with RoBERTa-small~\cite{liu2019roberta} as the decoder for text-line recognition, followed by LightGBM~\cite{ke2017lightgbm} for reading order.
In contrast, NDLkotenOCR-Lite uses RTMDet~\cite{lyu2022rtmdet} for layout recognition and PARSeq~\cite{bautista2022scene} for text-line recognition while retaining the reading-order method.

Furthermore, several datasets and competitions for Kuzushiji OCR, as well as studies focusing on specific OCR sub-tasks, have emerged in recent years.
For example, in 2019, several Japanese research institutions jointly organized the Kuzushiji Recognition competition on Kaggle and released an annotated dataset for the competition~\cite{kitamoto2020kaggle}.
In addition, Sakana AI introduced Metom~\cite{imajuku2024metom} in 2024, a Vision Transformer (ViT)-based~\cite{dosovitskiy2021an} classifier designed to recognize individual Kuzushiji characters.
The model was trained on 649,932 character samples, validated on 216,644 samples, and evaluated on 216,645 samples, achieving a micro accuracy of 97.22\%.
However, it operates at the individual-character level and does not directly process full historical document images.


\section{Proposed Method}
\subsection{Overall Framework}
This work proposes a multi-stage KCR framework that integrates a character-level Kuzushiji OCR pipeline with LLM-based post-OCR correction, as shown in Figure~\ref{fig:pipeline}.
The OCR pipeline sequentially performs character detection and cropping, Unicode-based character classification, and character ordering to produce an initial OCR result.
LLMs accessed via APIs are then employed to correct errors in the initial OCR result and produce the final OCR result.
In addition, we investigated a document restoration stage after character detection.
However, because it did not consistently improve performance across the test sets, it is not included in the main framework; details of this stage and experimental results are provided in Appendix~\ref{sec:restoration}.

\subsection{Synthetic Data Augmentation}
To enhance the robustness of detection models against seal interference, we propose a synthetic data augmentation strategy that constructs synthetic samples by randomly overlaying red seals onto original page images.
Specifically, the page images in the original training set contain at most two real seals, providing insufficient examples of complex seal-character overlap.
Therefore, synthetic samples with severe seal interference are generated using collected high-resolution seal images.

These seal images were collected from pre-modern East Asian historical artifacts and primarily contain Chinese characters and traditional decorative patterns.
The synthetic samples provide additional cases of seal-character overlap during training, enabling the detection model to detect characters more robustly under seal interference.

For the synthesis process, the seal images are first processed to remove their backgrounds and then randomly scaled so that their longest side ranges from 100 to 300 pixels, simulating seals of varying sizes in real historical documents.
Since page layouts vary across images, the valid page region is determined individually for each page image based on its layout.
To prevent seals from extending beyond this region, each seal is placed entirely within the valid page region.
When placing each synthetic seal, seal overlap is constrained such that no more than two seals overlap at any given location.
Finally, the scaled transparent seals are overlaid onto the original page images using alpha compositing.
As shown in Table~\ref{tab:sda}, after applying the proposed SDA strategy, the sizes of the training and validation sets increased from 800 to 1,600 page images and from 100 to 200 page images, respectively.
Overall, approximately 10 synthetic seals are overlaid on each synthetic image.

\begin{table}[t]
\centering
\caption{Statistics of the dataset splits and the proposed SDA strategy, where S/P denotes seals per page.}
\setlength{\tabcolsep}{4pt}
\begin{tabular}{@{}lrrrrr@{}}
\toprule
\multirow{2}{*}{\textbf{Set}} & \multicolumn{2}{c}{\textbf{Real Pages}} &
\multicolumn{2}{c}{\textbf{Synth. Pages}} & \multirow{2}{*}{\begin{tabular}[c]{@{}c@{}}\textbf{Total}\\ \textbf{Pages}\end{tabular}} \\ 
\cmidrule(lr){2-3} \cmidrule(lr){4-5}
& \textbf{\#Pages} & \textbf{S/P} & \textbf{\#Pages} & \textbf{S/P} & \\ 
\midrule
Train & 800 & $\leq 2$ & 800 & $\approx$10 & 1,600 \\
Validation & 100 & $\leq 2$ & 100 & $\approx$10 & 200 \\
Real Test & 100 & $\leq 2$ & -- & -- & 100 \\
Synth. Test & -- & -- & 100 & $\approx$10 & 100 \\ 
OOD Test & 1,090 & $\leq 2$ & -- & -- & 1,090 \\
\bottomrule
\end{tabular}
\label{tab:sda}
\end{table}

\begin{figure}[t]
\centering
\includegraphics[width=\linewidth]{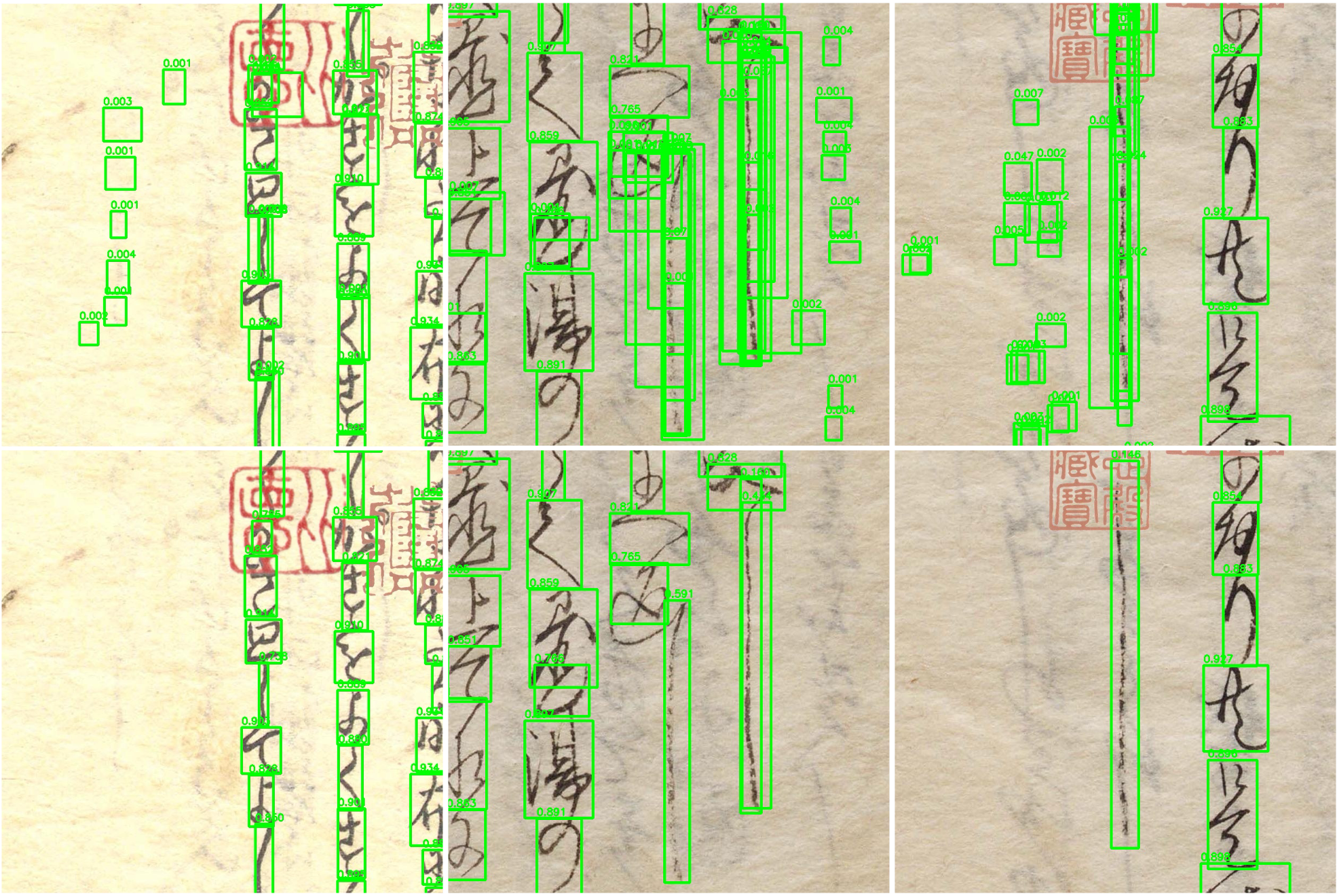}
\caption{Effect of confidence thresholding on detection.}
\label{fig:confi}
\end{figure}

\begin{figure}[t]
\centering
\includegraphics[width=\linewidth]{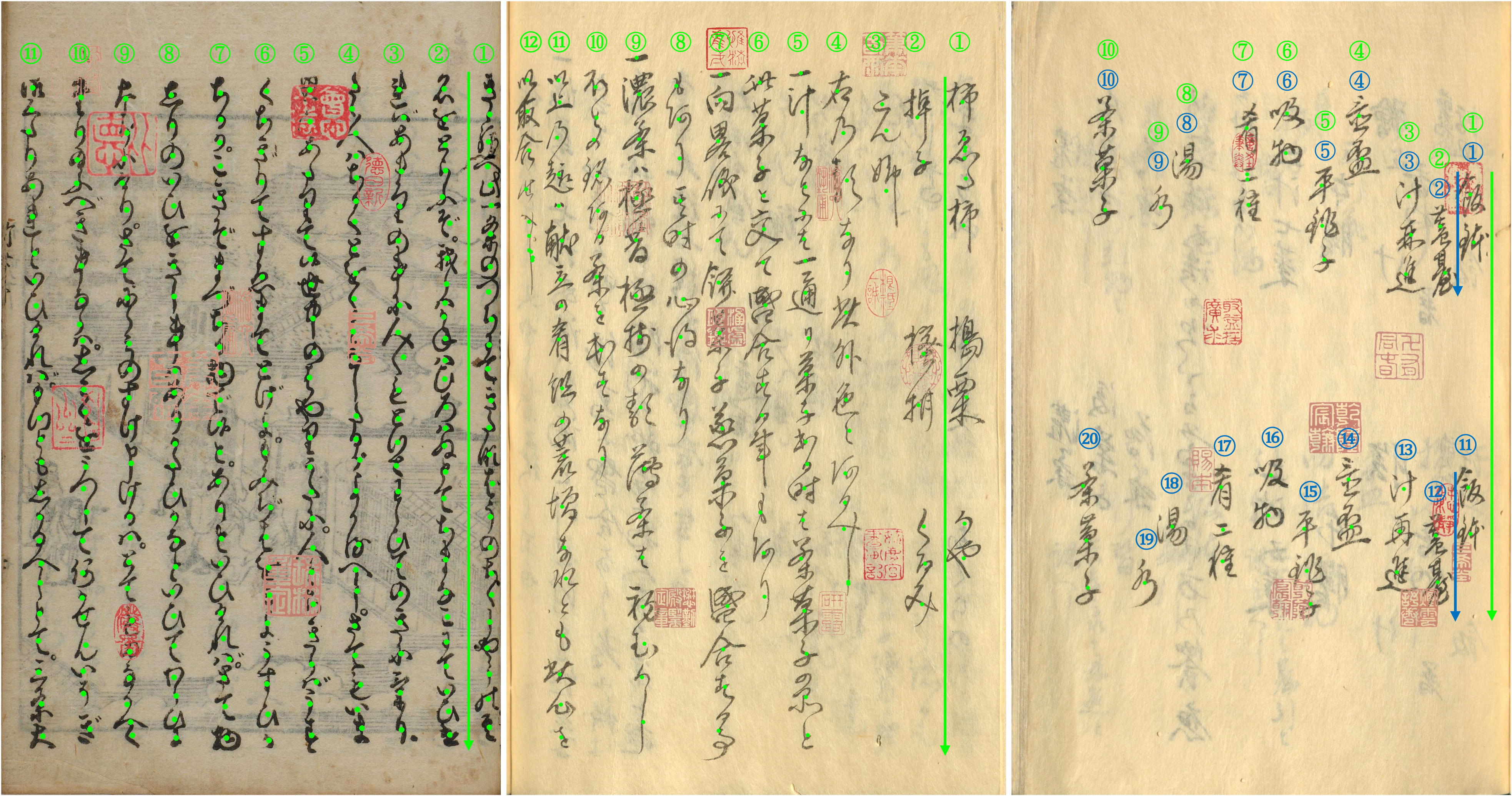}
\caption{Examples of character ordering with AdaCC, with green annotations indicating the AdaCC ordering.
The two left examples show correct ordering in regular layouts, whereas the right example shows an ordering in a complex layout, with the correct order shown in blue.}
\label{fig:ordering}
\end{figure}

\subsection{Character Detection and Cropping}
This work employs object detection models to detect Kuzushiji characters in full-page images of Japanese historical documents.
Under seal interference, character detection becomes challenging because seals often contain engraved characters visually similar to Kuzushiji characters, which may be mistakenly detected as Kuzushiji characters.
Dense character arrangements and complex character styles pose challenges, including missed detections, duplicate detections of partial character regions, and false positives caused by document stains.
Specifically, stains may be mistakenly detected as characters with extremely low confidence scores (e.g., 0.001), as shown in Figure~\ref{fig:confi}.
Furthermore, characters with elongated strokes or highly cursive forms may be detected multiple times as separate instances, as shown by the two rightmost examples in Figure~\ref{fig:confi}.
These duplicate detections, which typically correspond to partial character regions, generally have confidence scores below 0.1.
Thus, only bounding boxes with confidence scores greater than or equal to 0.1 are retained to suppress low-confidence false and duplicate detections.

After character detection, the top-left coordinates $(x, y)$, width $w$, and height $h$ of each predicted bounding box are used to crop the corresponding character region, which then serves as input for subsequent character classification.

\subsection{Character Classification}
Character classification is performed for each cropped character patch.
Given the complex character styles and diverse character categories, we employ the pre-trained Metom model~\cite{imajuku2024metom} as the character classifier, which supports 2,703 character categories.
Metom is a single-character classifier that assigns each character patch to one of the predefined character classes associated with Unicode code points and returns the corresponding Unicode character label.
To improve inference efficiency, character patches are processed in batches rather than individually.
In addition, the Scaled Dot-Product Attention (SDPA)~\cite{vaswani2017attention} implementation is enabled to improve inference throughput.
For evaluation, the top-5 predictions are recorded for each character patch, while the top-1 prediction for each patch is used as the final classification result.

\begin{algorithm}[t]
\caption{Adaptive Column Clustering.}
\fontsize{10pt}{13pt}\selectfont
\begin{algorithmic}[1]
\REQUIRE Predictions $\{x_i, y_i, w_i, h_i, \hat{c}_i\}_{i=1}^{N}$, $\lambda = 0.8$
\ENSURE Ordered Text $T$
\STATE $\mathcal{C}\gets [\;],\quad cx_i\gets x_i+\frac{w_i}{2},\quad cy_i\gets y_i+\frac{h_i}{2}$
\STATE $\mathcal{R}\gets \mathrm{sort}(\{1,\ldots,N\}, cx_i, \downarrow)$
\STATE $\tau \gets \lambda \cdot \frac{1}{N}\sum_{i=1}^{N}w_i$
\FOR{$i \in \mathcal{R}$}
\IF{$\mathcal{C}=\emptyset$}
\STATE $\mathcal{C}\gets \mathcal{C}\mathbin{+\!\!+}[\{i\}],\quad cx_1^{col} \gets cx_i$
\ELSE
\STATE $j^*\gets\arg\min_{1\leq j\leq |\mathcal{C}|} |cx_i-cx_j^{col}|$
\IF{$|cx_i-cx_{j^*}^{col}|\leq \tau$}
\STATE $\mathcal{C}_{j^*}\gets \mathcal{C}_{j^*}\cup\{i\}$
\STATE $cx_{j^*}^{col}\gets \mathrm{median}(\{cx_k\mid k\in\mathcal{C}_{j^*}\})$
\ELSE
\STATE $\mathcal{C}\gets \mathcal{C}\mathbin{+\!\!+}[\{i\}],\quad cx_{|\mathcal{C}|}^{col} \gets cx_i$
\ENDIF
\ENDIF
\ENDFOR
\STATE $T\gets \mathrm{Concat}_{\mathcal{C}_j\in\mathcal{C}} 
\Big( \hat{c}_k \mid k\in \mathrm{sort}(\mathcal{C}_j, cy_k, \uparrow) \Big)$
\RETURN $T$
\end{algorithmic}
\label{alg:ordering}
\end{algorithm}

\subsection{Character Ordering}
We propose a training-free character ordering algorithm, termed Adaptive Column Clustering (AdaCC), as detailed in Algorithm~\ref{alg:ordering}, with representative examples shown in Figure~\ref{fig:ordering}.
The algorithm takes the predicted bounding box parameters of each character as input, including $x_i$, $y_i$, $w_i$, and $h_i$, while the predicted character label $\hat{c}_i$ is used only to generate the ordered text output $T$.

The algorithm first initializes the spatial information required for column clustering.
Specifically, the column list $\mathcal{C}$ is initialized, and the center coordinates $(cx_i, cy_i)$ of each character are computed from its bounding box.
The column center $cx_j^{col}$ represents the horizontal position of the $j$-th column.
The characters are then sorted in descending order of $cx_i$ and processed from right to left.

To determine whether a character belongs to an existing column, the threshold $\tau$ is defined as $\lambda$ times the average predicted character width, where $\lambda = 0.8$.
If no column has yet been established, a new column is created and its center is initialized to $cx_i$.
Otherwise, the algorithm identifies the column whose center is closest to the current character along the x-axis and computes the corresponding horizontal distance.
If the distance does not exceed $\tau$, the character is assigned to that column, and its center $cx_j^{col}$ is updated as the median of the horizontal center coordinates of all characters assigned to the column.
Otherwise, a new column is appended to $\mathcal{C}$, with its center initialized to $cx_i$.

After all characters have been assigned to columns, the characters within each column are sorted in ascending order of $cy_i$, thereby arranging them from top to bottom.
Because new columns are appended to $\mathcal{C}$ while characters are processed in descending order of $cx_i$, the column list naturally preserves the right-to-left reading order without additional column sorting.
Finally, the predicted character labels $\hat{c}_i$ are concatenated according to the column order and the character order within each column to produce the ordered text output $T$.

\subsection{Post-OCR Correction}
Fine-tuning LLMs typically requires substantial amounts of high-quality task-specific data, while publicly available corpora of Japanese historical documents remain limited.
Therefore, we employ state-of-the-art (SOTA) proprietary LLMs accessed via APIs to refine Kuzushiji OCR outputs.
This stage operates solely on the recognized text and requires neither modification nor retraining of the preceding OCR stages.
Let $T_{\mathrm{OCR}}$ denote the OCR output and $P$ the Japanese correction prompt.
The final corrected output $\hat{T}$ is obtained as
\begin{equation}
\hat{T}=G(T_{\mathrm{OCR}};P),
\end{equation}
where $G(\cdot)$ denotes a proprietary LLM accessed via an API.
The model is instructed to correct OCR errors based on textual context while preserving historical kana orthography, traditional and variant character forms, and Classical Chinese-style expressions, without introducing unsupported completions or modernization.
The same prompt is used across all LLMs for consistent comparison, with details provided in Appendix~\ref{sec:prompt}.


\begin{table*}[t]
\centering
\caption{Quantitative comparison of character detectors under three data configurations on both test sets. 
Within each detector group, the best result is shown in \textbf{bold}, and the best overall result across all detectors is \underline{underlined}.}
\setlength{\tabcolsep}{5.2pt}
\begin{tabular}{@{}lcccccccccc@{}}
\toprule
\multirow{2}{*}{\textbf{Method}} & \multicolumn{2}{c}{\textbf{Model Complexity}} & \multicolumn{4}{c}{\textbf{Real Test Set}} & \multicolumn{4}{c}{\textbf{Synthetic Test Set}} \\ 
\cmidrule(lr){2-3}\cmidrule(lr){4-7}\cmidrule(lr){8-11}
& \textbf{Params} & \textbf{FLOPs} & \textbf{P} & \textbf{R} & $\textbf{AP}_{\textbf{50}}$ & $\textbf{AP}_{\textbf{50:95}}$ & \textbf{P} & \textbf{R} & $\textbf{AP}_{\textbf{50}}$ & $\textbf{AP}_{\textbf{50:95}}$ \\ 
\midrule
RT-DETR-R50 & 41.94M & 125.6G & 93.6 & 90.3 & 94.3 & 70.1 & 91.1 & 85.0 & 90.8 & 65.2 \\
+ Data Duplication & 41.94M & 125.6G & 95.8 & 92.2 & 95.1 & 69.5 & 93.7 & 87.0 & 92.1 & 64.8 \\
+ SDA (Ours) & 41.94M & 125.6G & \textbf{96.3} & \textbf{93.1} & \textbf{95.6} & \textbf{70.5} & \textbf{95.6} & \textbf{92.2} & \textbf{95.0} & \textbf{68.4} \\ 
\midrule
YOLOv9-C & 25.32M & 102.3G & \textbf{97.8} & 93.3 & \underline{\textbf{96.5}} & 82.6 & 95.8 & 88.0 & 93.8 & 77.8 \\
+ Data Duplication & 25.32M & 102.3G & 97.6 & 93.5 & \underline{\textbf{96.5}} & 82.8 & 95.6 & 88.4 & 94.0 & 78.1 \\
+ SDA (Ours) & 25.32M & 102.3G & \textbf{97.8} & \underline{\textbf{93.7}} & \underline{\textbf{96.5}} & \textbf{82.9} & \textbf{97.7} & \underline{\textbf{93.1}} & \underline{\textbf{96.4}} & \textbf{81.7} \\ 
\midrule
YOLOv10-L & 25.77M & 127.2G & 98.0 & 92.0 & 96.3 & 82.5 & 95.7 & 87.2 & 93.6 & 77.5 \\
+ Data Duplication & 25.77M & 127.2G & \underline{\textbf{98.4}} & 92.6 & \underline{\textbf{96.5}} & \textbf{83.1} & 96.3 & 87.9 & 94.1 & 78.3 \\
+ SDA (Ours) & 25.77M & 127.2G & 98.3 & \textbf{92.7} & \underline{\textbf{96.5}} & \textbf{83.1} & \underline{\textbf{98.0}} & \textbf{92.4} & \underline{\textbf{96.4}} & \textbf{81.8} \\ 
\midrule
YOLO11-L & 25.28M & 86.6G & 97.9 & 92.9 & 96.4 & 83.0 & 95.8 & 87.7 & 93.7 & 78.1 \\
+ Data Duplication & 25.28M & 86.6G & 97.7 & \textbf{93.6} & \underline{\textbf{96.5}} & 83.0 & 95.4 & 88.4 & 93.8 & 78.3 \\
+ SDA (Ours) & 25.28M & 86.6G & \textbf{98.0} & 93.5 & \underline{\textbf{96.5}} & \underline{\textbf{83.3}} & \textbf{97.6} & \underline{\textbf{93.1}} & \underline{\textbf{96.4}} & \underline{\textbf{82.1}} \\ 
\midrule
YOLOv12-L & 26.39M & 82.1G & 97.4 & 93.4 & 96.4 & 82.7 & 95.5 & 87.7 & 93.7 & 77.6 \\
+ Data Duplication & 26.39M & 82.1G & \textbf{97.9} & 93.4 & \underline{\textbf{96.5}} & \textbf{83.2} & 95.5 & 87.9 & 93.7 & 78.2 \\
+ SDA (Ours) & 26.39M & 82.1G & 97.5 & \textbf{93.6} & \underline{\textbf{96.5}} & \textbf{83.2} & \textbf{97.5} & \textbf{92.8} & \textbf{96.3} & \textbf{81.9} \\ 
\bottomrule
\end{tabular}
\label{tab:detection}
\end{table*}

\section{Experiments}
\subsection{Dataset Preparation}
We selected 1,000 publicly available page images from 13 Japanese historical books~\cite{genjimonogatari}.
The sources and statistics of these images are provided in Appendix~\ref{sec:dataset_source}.
We then systematically examined the annotations of all 1,000 images and identified missing main-text character annotations in 267 images.
The missing main-text character annotations were manually added with the assistance of a Kuzushiji expert.
Examples of these annotation corrections are provided in Appendix~\ref{sec:dataset_correction}.
After correcting the annotations, all images were randomly divided at the page level into training, validation, and test sets in an 8:1:1 ratio, resulting in 800, 100, and 100 images, respectively.

\subsection{Evaluation Sets}
We refer to the 100-image test split as the Real Test Set.
Following the seal synthesis strategy described in~\cite{ju2026dkds}, we constructed a Synthetic Test Set by overlaying seals on the 100 page images in the Real Test Set to simulate severe seal interference.
The Real and Synthetic Test Sets therefore contain 100 images each, for a total of 200 evaluation images.
Examples of page images before and after seal synthesis are provided in Appendix~\ref{sec:synthetic_test}.

Because the page-level random split results in pages from the same book appearing in different subsets, the testing data may share similar character styles, page layouts, and paper conditions with the training data.
To further evaluate generalization beyond the books used for training, we selected an additional 1,090 page images from seven Japanese historical books disjoint from the above 13 books, forming the out-of-domain (OOD) Test Set.
The sources and statistics of these seven books are provided in Appendix~\ref{sec:ood_dataset}.
Overall, the dataset splits and evaluation sets are summarized in Table~\ref{tab:sda}.

\subsection{Evaluation Metrics}
We conduct both stage-wise evaluations and an ablation study of the proposed KCR framework.
For character detection, we employ standard object detection metrics~\cite{lin2014microsoft}, including precision (P), recall (R), average precision at an IoU threshold of 0.50 ($\mathrm{AP}_{50}$), and average precision averaged over IoU thresholds from 0.50 to 0.95 ($\mathrm{AP}_{50:95}$).
We also report the number of model parameters (Params) and floating-point operations (FLOPs) to assess model complexity.
For character classification, we adopt Top-1 and Top-5 accuracy~\cite{krizhevsky2012imagenet}.
For character ordering, post-OCR correction, and the ablation study, we use micro-averaged character error rate (CER) computed over each test set.
P, R, AP, accuracy, and CER are reported as percentages (\%).
Overall, higher values indicate better performance for P, R, AP, and accuracy, whereas lower CER indicates better recognition performance; lower Params and FLOPs indicate lower model complexity.

\subsection{Implementation Details}
All local experiments were conducted on an Ubuntu system equipped with an Intel Core i5-12600K CPU and an NVIDIA RTX PRO 6000 Blackwell GPU.
For inference, character detection and classification were performed on the GPU, character cropping and ordering were performed on the CPU, and post-OCR correction was performed via API.

For character detection, all models were trained for 1,000 epochs with a batch size of 16 and an input size of $640 \times 640$ pixels.
Stochastic gradient descent (SGD) was used as the optimizer with a learning rate of 0.01.
The checkpoint achieving the best validation performance was selected for evaluation.
For character ordering, the threshold scale $\lambda$ and column-center update strategy were determined experimentally.
We set $\lambda=0.8$ and used the median for column-center updates.
For post-OCR correction, the maximum number of output tokens was set to 1,024.
The reasoning effort was set to ``none'' for the GPT-5 series and ``low'' for GPT-6-Astra.
For the GPT-5 series, the temperature was set to 0 and top-$p$ to 1.
Each LLM was independently evaluated three times, and the mean and standard deviation were reported.

\subsection{Character Detection}\label{sec:detection}
We evaluated several object detectors, including RT-DETR~\cite{zhao2024detrs}, YOLOv9~\cite{wang2024yolov9}, YOLOv10~\cite{wang2024yolov10}, YOLO11~\cite{jocher2024yolo11}, and YOLOv12~\cite{tian2026yolov12}, using three data configurations: the original configuration, Data Duplication, and the proposed SDA strategy.
Data Duplication simply duplicates the original training data and serves as a control for distinguishing the effect of increased training-set size from that of SDA.
Under SDA, synthetic counterparts are added to both the training and validation sets, increasing them from 800 to 1,600 images and from 100 to 200 images, respectively.
As shown in Table~\ref{tab:detection}, Data Duplication and SDA alter only the data configuration and do not modify the detector architecture; therefore, Params and FLOPs remain unchanged for each detector.

On the Real Test Set, SDA remains competitive with or outperforms Data Duplication.
Its advantage is more pronounced on the Synthetic Test Set, where it consistently outperforms both alternatives across all evaluated detectors.
Compared with the original configuration, SDA improves recall by 5.1 to 7.2 percentage points, $\mathrm{AP}_{50}$ by 2.6 to 4.2 percentage points, and $\mathrm{AP}_{50:95}$ by 3.2 to 4.3 percentage points on the Synthetic Test Set.
By contrast, Data Duplication yields only marginal and inconsistent gains.
These results suggest that the improvements achieved by SDA arise primarily from exposure to synthetic seal interference rather than from the increased training-set size alone.

Among all evaluated configurations, YOLO11-L trained with SDA achieves the highest $\mathrm{AP}_{50:95}$ on both the Real and Synthetic Test Sets, reaching 83.3\% and 82.1\%, respectively, while maintaining strong performance across the other detection metrics.
We therefore adopt YOLO11-L trained with SDA as the detector in the proposed framework.

\begin{table}[t]
\centering
\caption{Character classification performance of our framework on the Real and Synthetic Test Sets.}
\setlength{\tabcolsep}{20.5pt}
\begin{tabular}{@{}lcc@{}}
\toprule
\textbf{Test Set} & \textbf{Top-1 Acc.} & \textbf{Top-5 Acc.} \\
\midrule
Real & 97.40\% & 99.38\% \\
Synth. & 93.93\% & 97.48\% \\
\bottomrule
\end{tabular}
\label{tab:classification}
\end{table}

\begin{table}[t]
\centering
\caption{Post-OCR CER (\%) of different LLMs over three runs, reported as mean $\pm$ standard deviation.}
\setlength{\tabcolsep}{8pt}
\begin{tabular}{@{}lcc@{}}
\toprule
\textbf{Model} & \textbf{Real Test Set} & \textbf{Synth. Test Set} \\
\midrule
GPT-5.4      & $14.02 \pm 0.18$ & $18.56 \pm 0.03$ \\
GPT-5.5      & $11.22 \pm 0.03$ & $14.80 \pm 0.05$ \\
GPT-5.6-Sol  & $14.34 \pm 0.02$ & $18.53 \pm 0.42$ \\
GPT-6-Astra  & $11.55 \pm 0.03$ & $14.83 \pm 0.09$ \\
\bottomrule
\end{tabular}
\label{tab:postocr}
\end{table}

\subsection{Character Classification}
After character detection using YOLO11-L~\cite{jocher2024yolo11} trained with SDA, we cropped the predicted bounding boxes and classified the resulting character patches using Metom~\cite{imajuku2024metom}.
For classification evaluation, detected character regions were associated with their ground-truth character annotations.
Table~\ref{tab:classification} presents the classification results on the Real and Synthetic Test Sets.
The classifier achieves high recognition accuracy, with Top-1 accuracies of 97.40\% and 93.93\% on the Real and Synthetic Test Sets, respectively.
The lower Top-1 accuracy on the Synthetic Test Set indicates that severe seal interference makes classification more challenging.
Nevertheless, the corresponding Top-5 accuracies remain high at 99.38\% and 97.48\%, suggesting that the correct character often remains among the top-ranked candidates even under such interference.

\subsection{Post-OCR Correction}
In the post-OCR correction stage, we evaluated four LLMs, including GPT-5.4~\cite{openai2026gpt54}, GPT-5.5~\cite{openai2026gpt55}, GPT-5.6-Sol~\cite{openai2026gpt56}, and GPT-6-Astra~\cite{openai2026gpt6}, on the Real and Synthetic Test Sets.
For all LLMs, the input was the OCR output produced by YOLO11-L with SDA, Metom, and AdaCC.
As shown in Table~\ref{tab:postocr}, GPT-5.5 achieved the lowest CERs on both test sets, at 11.22\% and 14.80\% for the Real and Synthetic Test Sets, respectively.
GPT-6-Astra showed comparable performance, with CERs of 11.55\% and 14.83\%, respectively.
In contrast, GPT-5.4 and GPT-5.6-Sol yielded substantially higher CERs, particularly on the Synthetic Test Set, where both exceeded 18\%.
All models exhibited higher CERs on the Synthetic Test Set than on the Real Test Set, indicating that errors induced by severe synthetic seal interference remain challenging even after post-OCR correction.
Accordingly, GPT-5.5 was selected for the final framework configuration.

\begin{table*}[t]
\centering
\caption{Ablation study of the proposed components and comparison with existing systems in terms of CER (\%).}
\setlength{\tabcolsep}{15pt}
\begin{tabular}{@{}lcccccc@{}}
\toprule
\multirow{2}{*}{\textbf{Method}} & \multicolumn{3}{c}{\textbf{Component}} & \multicolumn{3}{c}{\textbf{Performance (CER)}} \\
\cmidrule(lr){2-4} \cmidrule(l){5-7}
& \textbf{SDA} & \textbf{AdaCC} & \textbf{Post-OCR} & \textbf{Real} & \textbf{Synth.} & \textbf{OOD} \\
\midrule
\multirow{5}{*}{Our Framework}
&            &            &            & 19.85 & 27.42 & 20.66 \\
& \checkmark &            &            & 19.30 & 22.78 & 19.74 \\
&            & \checkmark &            & 12.01 & 20.26 & 14.28 \\
& \checkmark & \checkmark &            & 11.59 & 15.29 & 13.42 \\
& \checkmark & \checkmark & \checkmark & \textbf{11.22} & \textbf{14.80} & \textbf{12.58} \\
\midrule
NDLkotenOCR-Lite & -- & -- & -- & 21.76 & 47.82 & 27.84 \\
NDLkotenOCR      & -- & -- & -- & 12.73 & 23.43 & 22.64 \\
\bottomrule
\end{tabular}
\label{tab:overall}
\end{table*}

\subsection{Ablation Study}
To evaluate the effectiveness of the proposed framework, we conducted an ablation study of its components and compared the proposed framework with existing Kuzushiji OCR systems.
Notably, for character ordering, we adopted the LightGBM-based reading-order method~\cite{ke2017lightgbm} used in Version 1 of NDLkotenOCR~\cite{kiyonori2023enhancing} as the supervised baseline and adapted it to character-level bounding boxes.
Since LightGBM requires supervised training, we specifically trained it using the training split of the benchmark dataset.

As shown in Table~\ref{tab:overall}, the baseline without any proposed components achieved CERs of 19.85\%, 27.42\%, and 20.66\% on the Real, Synthetic, and OOD Test Sets, respectively.
Introducing SDA reduced them to 19.30\%, 22.78\%, and 19.74\%, with the largest improvement on the Synthetic Test Set, indicating increased robustness to severe seal interference.
Applying AdaCC independently further reduced the CERs to 12.01\%, 20.26\%, and 14.28\%, while combining SDA and AdaCC achieved 11.59\%, 15.29\%, and 13.42\%.
With post-OCR correction, the CERs were further reduced to 11.22\%, 14.80\%, and 12.58\%, yielding the best performance across all three test sets.
The consistent improvements on the OOD Test Set further indicate that the proposed components generalize to books unseen during training.

We further compared our framework with NDLkotenOCR~\cite{kiyonori2023enhancing} and NDLkotenOCR-Lite~\cite{toru2024development}, two existing Kuzushiji OCR systems with publicly available implementations.
We used the publicly released Version 3 model of NDLkotenOCR, trained on 28,134 images and 456,746 text lines, and the pre-trained NDLkotenOCR-Lite model, trained on 502,065 text lines.
On the Real Test Set, our framework achieved a CER of 11.22\%, compared with 12.73\% for NDLkotenOCR and 21.76\% for NDLkotenOCR-Lite.
On the Synthetic Test Set, it reduced the CER by 36.83\% and 69.05\% relative to NDLkotenOCR (23.43\%) and NDLkotenOCR-Lite (47.82\%), respectively.
On the OOD Test Set, the corresponding relative reductions were 44.43\% and 54.81\% compared with NDLkotenOCR (22.64\%) and NDLkotenOCR-Lite (27.84\%).
Overall, these results demonstrate substantially improved robustness to severe seal interference and stronger generalization to books unseen during training.

\subsection{Visualization}
\begin{figure}[t]
\centering
\includegraphics[width=\linewidth]{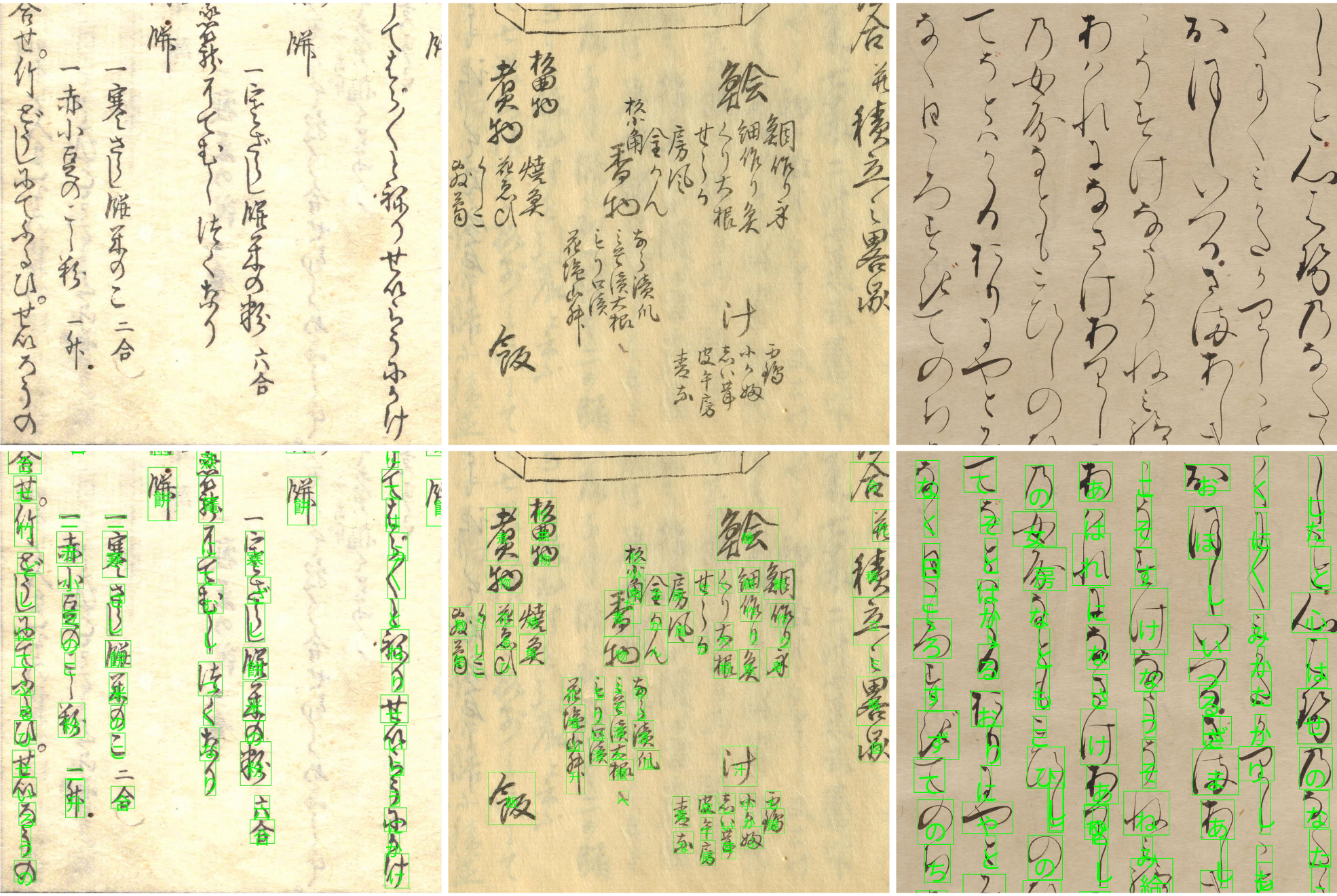}
\caption{Visualization of the input images (top) and recognition results overlaid on the corresponding images (bottom).
The font size is uniformly set to 64 pixels.}
\label{fig:visualization}
\end{figure}

One source of residual recognition errors is the presence of complex page layouts, in which the content of a single page is divided into multiple independent text blocks.
As shown in the right example of Figure~\ref{fig:ordering}, AdaCC performs character ordering under a regular-layout assumption, resulting in the ordering shown in green, whereas the page contains two independent text blocks that should be ordered separately, as shown in blue.
Such layouts occur in only a small portion of our evaluation data and exhibit limited structural consistency, making them difficult to process using simple rule-based methods.
Nevertheless, Figure~\ref{fig:visualization} suggests that the visualization results may still facilitate the interpretation of documents with complex layouts.

\section{Conclusion}
This work proposes a multi-stage KCR framework for Japanese historical documents.
Experiments on the Real, Synthetic, and OOD Test Sets show consistent improvements over the conventional character-level OCR baseline.
The results demonstrate the effectiveness of the proposed framework in improving recognition accuracy, robustness to seal interference, and generalization to Japanese historical books unseen during training.

Future work will focus on collecting more high-quality data, improving the processing of complex layouts, and deploying the framework through web and mobile applications for broader public use.

\section*{Limitations}
The experimental results for NDLkotenOCR and NDLkotenOCR-Lite were obtained using publicly released models pre-trained on substantially larger datasets, whereas the trainable components of our framework were optimized on the reconstructed, limited-scale training set used in this work.
Our framework also relies on the pre-trained Metom classifier and API-based LLMs whose original training data differ from those used by the compared OCR systems; in particular, the training data of the proprietary LLMs are not publicly available.
Thus, the compared systems were not trained under identical or fully controlled data conditions.

Moreover, NDLkotenOCR and NDLkotenOCR-Lite rely on text-line-level annotations, whereas our framework uses character-level annotations, leading to substantially different supervision requirements and training procedures.
These differences make it difficult to retrain all systems using the same data and supervision settings.
Therefore, the observed performance differences may reflect not only methodological differences but also variations in training-data scale, distribution, pre-training resources, and annotation granularity, which should be considered when interpreting the results.

\section*{Acknowledgments}
This work was supported by JSPS KAKENHI Grant Number 25H01242, and JST SPRING Grant Number JPMJSP2110.

\bibliography{ref}

\appendix

\begin{figure*}[t]
\centering
\includegraphics[width=\linewidth]{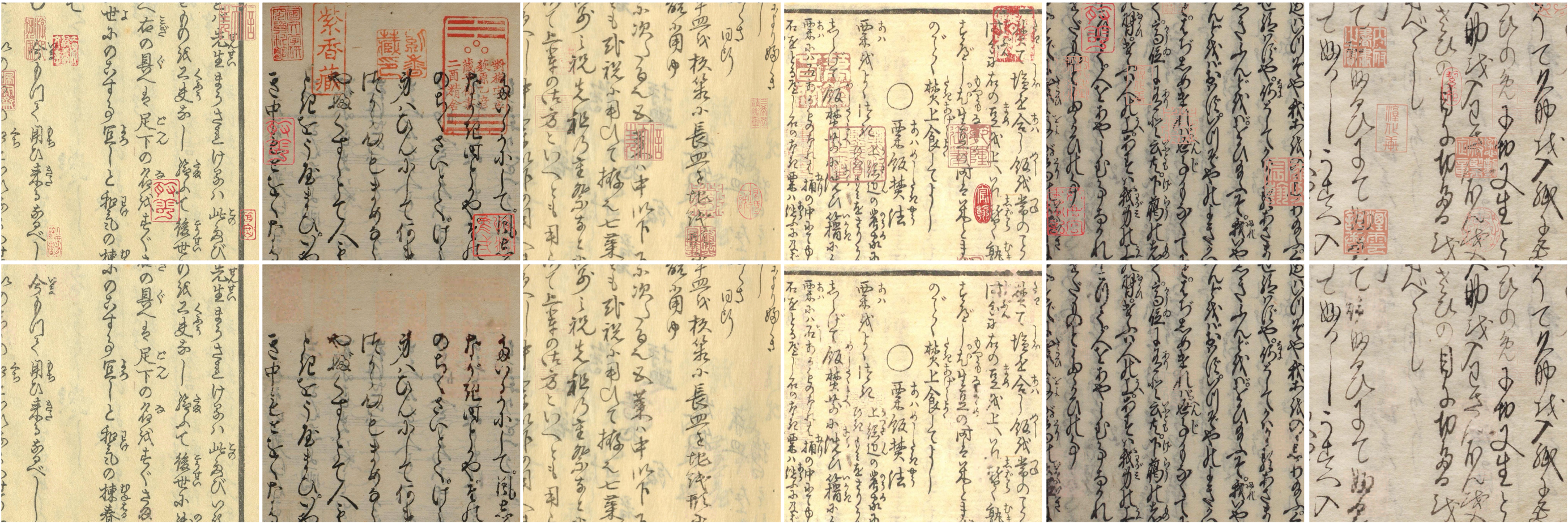}
\caption{Qualitative comparison of images from the Synthetic Test Set (top) and their restored images (bottom).}
\label{fig:restoration}
\vspace{-0.3em}
\end{figure*}

\section{Document Restoration}\label{sec:restoration}
\subsection{Restoration Method}
In addition to the components included in the main framework, we investigated a training-free, color-based thresholding method for document restoration.
Given an input page image $I \in \mathbb{R}^{H \times W \times 3}$, let $R$, $G$, and $B \in [0,255]$ denote the red, green, and blue channel intensities of each pixel.
Since red seal pixels typically exhibit higher red-channel intensities than green- and blue-channel intensities, they are detected using a channel-ratio-based thresholding rule. 
A pixel is classified as a red seal candidate if it satisfies:
\begin{equation}
(R \ge \tau_r) \wedge (R \ge \tau_{rg} \cdot G) \wedge (R \ge \tau_{rb} \cdot B),
\end{equation}
where $\tau_r$ denotes the minimum red-channel intensity threshold, while $\tau_{rg}$ and $\tau_{rb}$ control the dominance of the red channel over the green and blue channels, respectively.
This rule suppresses pixels without sufficient red-channel dominance while retaining pixels with strong red-channel responses.

The resulting binary mask $M$ is further refined using morphological dilation to slightly expand the detected seal boundaries and better compensate for color bleeding around the seal edges.
A $3 \times 3$ square kernel with one dilation iteration is used by default.
The detected seal regions are then removed through image inpainting:
\begin{equation}
I_{\text{restored}} = \text{Inpaint}(I, M, \rho),
\end{equation}
where $\rho$ denotes the inpainting radius.
We adopt the fast marching image inpainting method~\cite{telea2004image}, which efficiently reconstructs masked regions using information from surrounding pixels without requiring any additional training.

\begin{table}[t]
\centering
\caption{Parameter sensitivity analysis of $\tau_r$ and $(\tau_{rg}, \tau_{rb})$ for restoration on the validation and test sets.}
\setlength{\tabcolsep}{4pt}
\begin{tabular}{@{}lccccc@{}}
\toprule
\multirow{2}{*}{$\boldsymbol{\tau_r}$} & \multirow{2}{*}{$\boldsymbol{(\tau_{rg},\,\tau_{rb})}$} & \multicolumn{2}{c}{\textbf{Validation Set}} & \multicolumn{2}{c}{\textbf{Test Set}} \\ \cmidrule(lr){3-4}\cmidrule(lr){5-6}
& & \textbf{PSNR} & \textbf{SSIM} & \textbf{PSNR} & \textbf{SSIM} \\ \midrule
-- & -- & 29.15 & 0.9655 & 28.71 & 0.9639 \\ \midrule
80 & (1.2, 1.2) & 29.76 & 0.9470 & 29.61 & 0.9465 \\
80 & (1.3, 1.3) & 33.64 & 0.9736 & 33.73 & 0.9731 \\
80 & (1.4, 1.4) & 33.87 & 0.9756 & 33.77 & 0.9745 \\
80 & (1.5, 1.5) & 31.97 & 0.9717 & 31.68 & 0.9706 \\ \midrule
90 & (1.2, 1.2) & 30.37 & 0.9522 & 30.19 & 0.9619 \\
90 & (1.3, 1.3) & \textbf{34.09} & 0.9757 & \textbf{34.13} & 0.9750 \\
90 & (1.4, 1.4) & 34.05 & \textbf{0.9763} & 33.94 & \textbf{0.9753} \\
90 & (1.5, 1.5) & 32.03 & 0.9721 & 31.74 & 0.9710 \\ \bottomrule
\end{tabular}
\label{tab:restoration_sensitivity}
\end{table}

\begin{table}[t]
\centering
\caption{Effect of document restoration on KCR before post-OCR correction, measured by CER (\%).}
\setlength{\tabcolsep}{3pt}
\begin{tabular}{lcc}
\toprule
\textbf{Method} & \textbf{Real Test Set} & \textbf{Synth. Test Set} \\ 
\midrule
w/o Restoration & 11.59 & 15.29 \\
w/ Restoration  & 11.98 & 13.67 \\
\bottomrule
\end{tabular}
\label{tab:restoration_ablation}
\end{table}

\subsection{Parameter Sensitivity Analysis}
We evaluate restoration quality using peak signal-to-noise ratio (PSNR, in dB) and structural similarity index measure (SSIM)~\cite{wang2004image}.
The fast marching image inpainting method~\cite{telea2004image} is used with a fixed inpainting radius of $\rho=3$, while mask refinement uses a $3 \times 3$ dilation kernel with a single iteration.
We conduct a sensitivity analysis of the hyperparameters $\tau_r$ and $(\tau_{rg},\tau_{rb})$, with the results summarized in Table~\ref{tab:restoration_sensitivity}.

The first row represents the no-restoration baseline, in which the synthetically degraded images are directly compared with their corresponding original images.
Most restoration configurations improve PSNR over this baseline, although permissive channel-ratio thresholds such as $(\tau_{rg},\tau_{rb})=(1.2,1.2)$ result in lower SSIM scores.
Overall, $\tau_r=90$ performs better than $\tau_r=80$ under the same channel-ratio thresholds.
With $\tau_r=90$, $(\tau_{rg},\tau_{rb})=(1.3,1.3)$ achieves the highest PSNR of 34.09 dB on the Validation Set and 34.13 dB on the test set.
Increasing the channel-ratio thresholds to $(1.4,1.4)$ yields slightly higher SSIM scores of 0.9763 and 0.9753, respectively, but lower PSNR values.
Therefore, we select $\tau_r=90$ and $(\tau_{rg},\tau_{rb})=(1.3,1.3)$ for the experiments.

\subsection{Effect of Document Restoration}
We investigate the effect of document restoration on KCR performance before post-OCR correction.
As shown in Table~\ref{tab:restoration_ablation}, restoration increases the CER on the Real Test Set from 11.59\% to 11.98\%.
In contrast, it reduces the CER on the Synthetic Test Set from 15.29\% to 13.67\%, corresponding to a relative reduction of 10.60\%.
These results indicate that document restoration does not consistently improve KCR performance.
Although restoration can suppress seal interference and produce visually cleaner images, it may also alter fine-grained character strokes or introduce local artifacts that adversely affect character recognition.
We therefore treat document restoration as an optional component and exclude it from the main framework.

\begin{table*}[t]
\centering
\caption{Statistics of the benchmark dataset reconstructed from 13 Japanese historical books.
``Pages'' denotes the number of annotated page images, and ``Chars/Page'' denotes the average number of characters per page.}
\label{tab:dataset}
\setlength{\tabcolsep}{2.4pt}
\begin{tabular}{@{}lllrrr@{}}
\toprule
\textbf{Index} & \textbf{NIJL ID} & \textbf{Book Title} & \textbf{Pages} & \textbf{Chars.} & \textbf{Chars/Page} \\ \midrule
1  & 100241706 & Usonarubeshi (虚南留別志) & 67  & 8,565   & 127.8 \\
2  & 100249376 & Gozenkashi Hidenshō (御前菓子秘伝抄) & 104 & 11,841  & 113.9 \\
3  & 100249416 & Mochigashi Sokuseki Teseishū (餅菓子即席／手製集) & 58  & 7,967  & 137.4 \\
4  & 100249476 & Meshi Hyakuchinden (飯百珍伝) & 46  & 7,842   & 170.5 \\
5  & 200006663 & Jiguchi (ぢぐち) & 8   & 121     & 15.1 \\
6  & 200015843 & Nippon Eitaigura (日本永代蔵) & 180 & 50,251  & 279.2 \\
7  & 200017458 & Soga Monogatari (曾我物語) & 78  & 29,641  & 380.0 \\
8  & 200020019 & Chikusai (竹斎) & 146 & 33,228  & 227.6 \\
9  & 200021086 & Isoho Monogatari (伊曾保物語) & 60  & 15,410  & 256.8 \\
10 & 200021763 & Zenbu Ryōrishō (膳部料理抄) & 94  & 11,437  & 121.7 \\
11 & 200021802 & Ryōri Monogatari (料理物語) & 105 & 19,609  & 186.8 \\
12 & 200021869 & Ryōrikata Kokoroe no Koto (料理方心得之事) & 30 & 3,012 & 100.4 \\
13 & 200022050 & Ryōri Hidenshō (料理秘伝抄) & 24 & 9,558 & 398.3 \\ \midrule
Total & -- & -- & 1,000 & 208,482 & 208.5 \\ \bottomrule
\end{tabular}
\vspace{-0.4em}
\end{table*}

\begin{table*}[t]
\centering
\caption{Statistics of the OOD Test Set reconstructed from seven book-disjoint Japanese historical books.
``Pages'' denotes the number of annotated page images, and ``Chars/Page'' denotes the average number of characters per page.}
\label{tab:ood_dataset}
\setlength{\tabcolsep}{10pt}
\begin{tabular}{@{}lllrrr@{}}
\toprule
\textbf{Index} & \textbf{NIJL ID} & \textbf{Book Title} & \textbf{Pages} & \textbf{Chars.} & \textbf{Chars/Page} \\ \midrule
1 & 100249371 & Teisa Hiroku (鼎左秘録) & 83 & 9,580 & 115.4 \\
2 & 100249537 & Ryōri Chinmishū (料理珍味集) & 159 & 12,358 & 77.7 \\
3 & 200003076 & Kōshoku Ichidai Otoko (好色一代男) & 282 & 63,959 & 226.8 \\
4 & 200010454 & Genji Monogatari (源氏物語) & 71 & 11,566 & 162.9 \\
5 & 200014740 & Ugetsu Monogatari (雨月物語) & 154 & 44,832 & 291.1 \\
6 & 200015779 & Ukiyoburo (浮世風呂) & 268 & 60,381 & 225.3 \\
7 & 200018243 & Tamakushige (玉くしげ) & 73 & 13,623 & 186.6 \\ \midrule
Total & -- & -- & 1,090 & 216,299 & 198.4 \\ \bottomrule
\end{tabular}
\vspace{-0.4em}
\end{table*}

\subsection{Qualitative Evaluation}
Figure~\ref{fig:restoration} presents qualitative comparisons between synthetically seal-interfered input images and their corresponding restoration results using $\tau_r=90$ and $(\tau_{rg},\tau_{rb})=(1.3,1.3)$.
The input images contain severe seal interference, with many seals overlapping Kuzushiji characters.
After restoration, most red seal interference is effectively removed, although faint pink traces remain in some cases.

In regions with dense seal coverage, restoration may introduce locally brightened or over-smoothed areas and remove fine image details.
Such artifacts can damage local character stroke structures, which may explain why improved visual quality does not consistently translate into lower CER.

\section{Dataset Details}
\subsection{Benchmark Dataset Source}\label{sec:dataset_source}
The benchmark dataset was reconstructed from Kuzushiji data derived from digitized Japanese historical books held by the National Institute of Japanese Literature (NIJL)~\cite{genjimonogatari} and made available through the Center for Open Data in the Humanities (CODH).
The original dataset contains 6,151 page images with 1,086,326 annotated character instances spanning 4,328 character types.
Character-level annotations include bounding box coordinates and corresponding Unicode code points.
From this resource, we selected 13 Japanese historical books and excluded pages containing no annotated Kuzushiji characters.
The resulting collection consists of 1,000 page images.
After the annotation corrections described below, the benchmark dataset contains 208,482 Kuzushiji characters, with an average of 208.5 characters per page.
The bibliographic information and statistics of the 13 books are summarized in Table~\ref{tab:dataset}.

\begin{figure}[t]
\centering
\includegraphics[width=\linewidth]{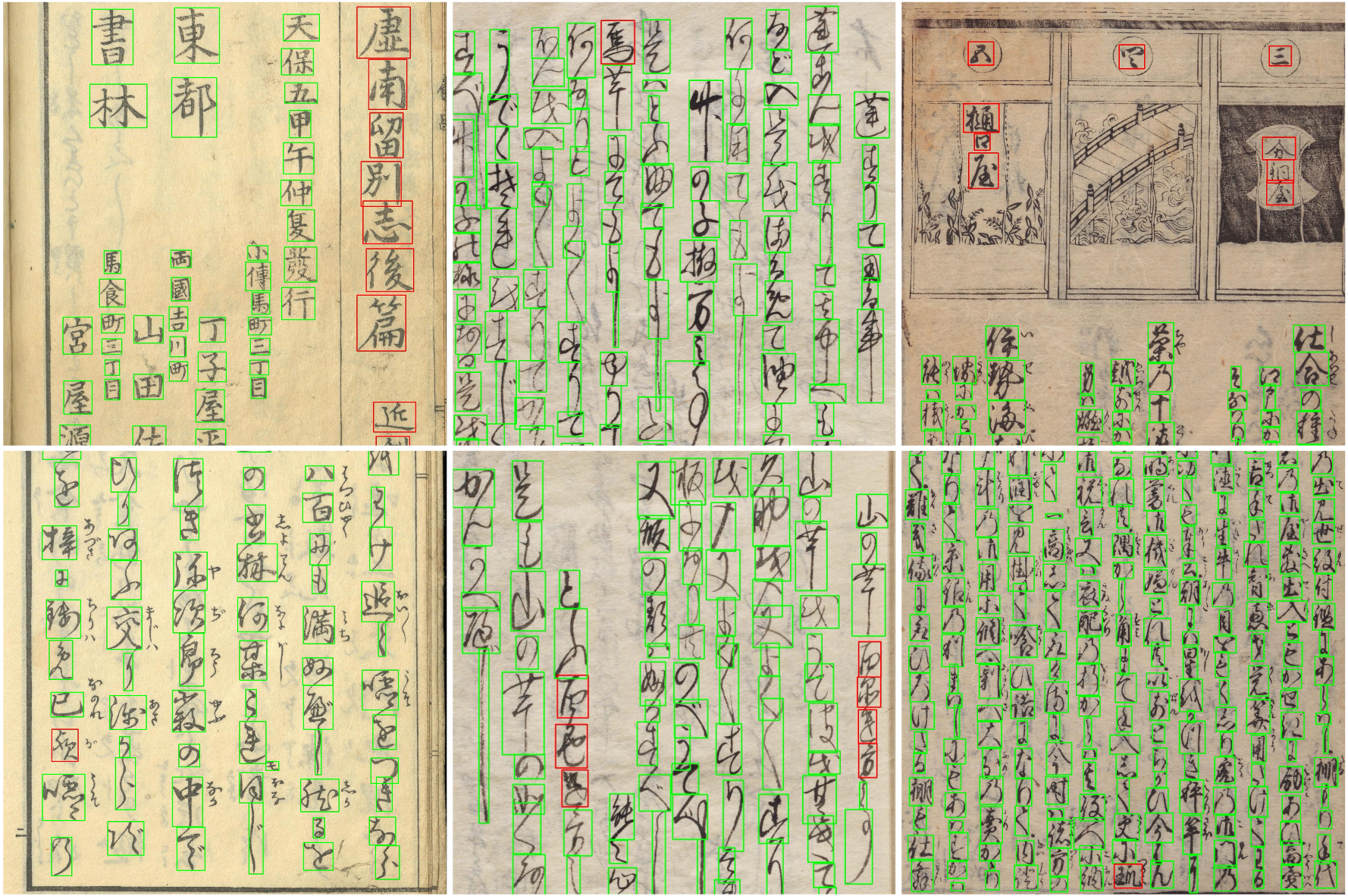}
\caption{Examples of annotation corrections.
Green bounding boxes indicate annotations originally included in the dataset, while red bounding boxes indicate missing character annotations added in this work.}
\label{fig:correction}
\end{figure}

\begin{figure}[t]
\centering
\includegraphics[width=\linewidth]{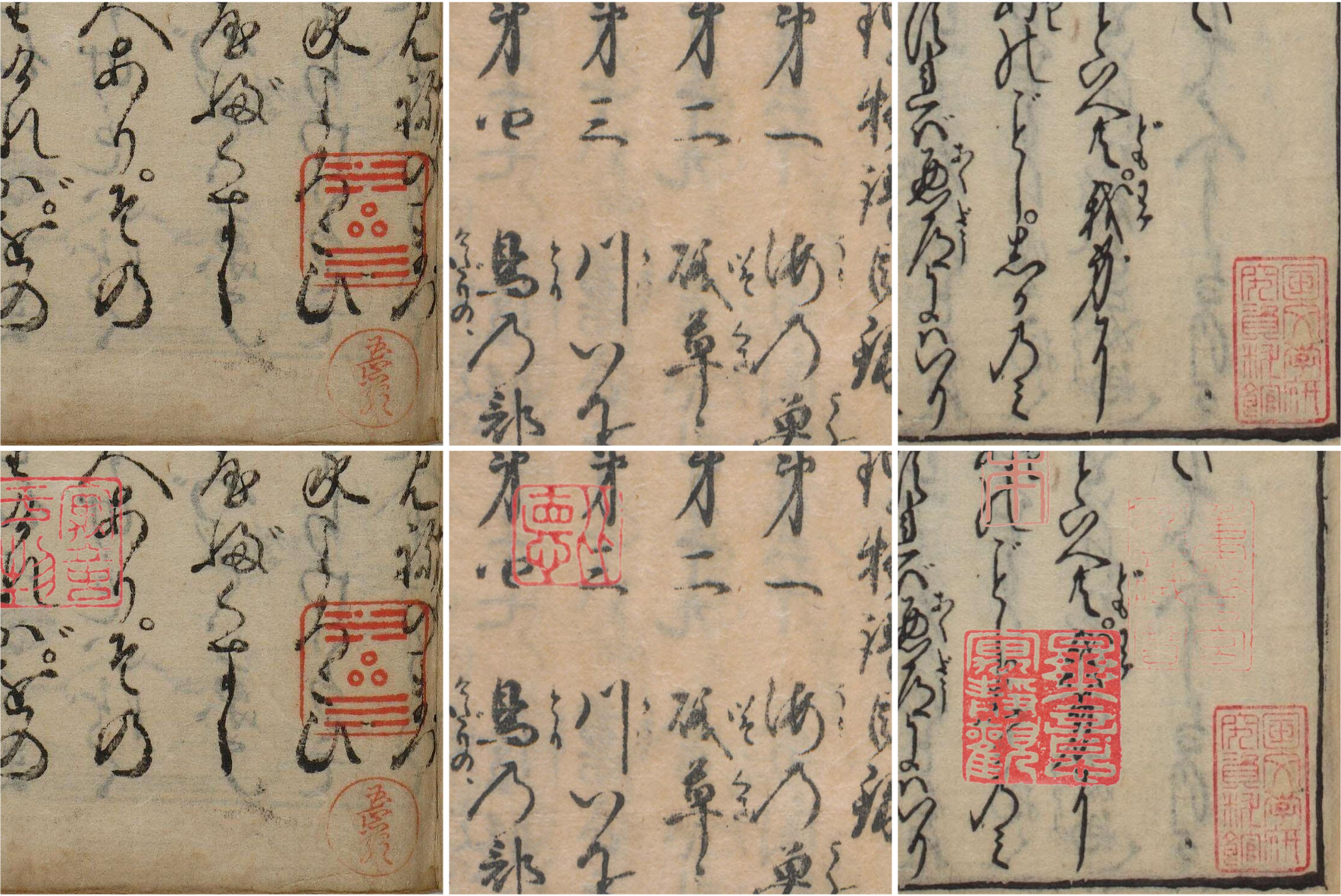}
\caption{Examples of real and synthetic seal interference.
The top row shows pages with naturally occurring seals, while the bottom row shows pages from the Synthetic Test Set with synthetically overlaid seals.}
\label{fig:synthetic}
\vspace{-1em}
\end{figure}

\subsection{Annotation Correction}\label{sec:dataset_correction}
We manually and systematically reviewed the character annotations of all 1,000 page images.
With the assistance of a Kuzushiji expert, annotation omissions were carefully examined to distinguish note text from characters belonging to the main text.
Following the original annotation rules, characters appearing in notes are intentionally excluded from annotation.
We therefore retained this annotation policy and specifically added annotations only for main-text Kuzushiji characters that had been inadvertently omitted.
Missing main-text annotations were identified in 267 images.
Figure~\ref{fig:correction} shows representative examples of these corrections.
Green bounding boxes denote the original annotations, while red bounding boxes indicate missing main-text character annotations added in this work.

\subsection{Synthetic Test Set}\label{sec:synthetic_test}
The Synthetic Test Set was constructed from the 100 page images in the Real Test Set following the seal synthesis strategy described in~\cite{ju2026dkds}.
We used high-resolution seal images distinct from those used for SDA during training.
Ten seal instances were overlaid on each page, with no more than two synthetic seals overlapping at any location.
The Real and Synthetic Test Sets therefore share the same underlying page images and character annotations, differing only in the added synthetic seal interference.
Figure~\ref{fig:synthetic} shows representative examples of real and synthetic seal interference.

\subsection{OOD Test Set Source}\label{sec:ood_dataset}
The OOD Test Set was also drawn from digitized historical books held by NIJL~\cite{genjimonogatari} but consists of seven historical books disjoint from the 13 books used to construct the benchmark dataset.
The bibliographic information and statistics of the seven books are summarized in Table~\ref{tab:ood_dataset}.
It contains 1,090 annotated page images with 216,299 Kuzushiji character instances, averaging 198.4 characters per page.
This book-disjoint setting enables evaluation of generalization to unseen books.

\section{Post-OCR Correction Prompt}\label{sec:prompt}
Because the source documents contain pre-modern Japanese, we used the same Japanese prompt for all LLMs to preserve linguistic nuances, historical orthography, and document-specific expressions while ensuring a consistent evaluation setting.
For each page, the prompt was provided as the system instruction, and the raw OCR output was supplied separately as input text.
The models were instructed to correct only obvious OCR errors while preserving the original content and character order, without introducing unsupported completions, summaries, explanations, or annotations.
The Japanese prompt used in this work is shown below, followed by an English translation for reference.

\smallskip
\noindent\textbf{Japanese prompt.}\par
\begin{promptbox}
あなたは日本古典籍OCRの誤り訂正システムです。

OCRによる明らかな誤字・脱字・衍字だけを、原文の文脈に基づいて訂正してください。
歴史的仮名遣い、旧字体、異体字、漢文調の表現は現代語に変更しないでください。
根拠のない補完、要約、説明、注釈を加えず、原文の内容と順序を維持してください。
修正後の本文のみを出力してください。
\end{promptbox}

\noindent\textbf{English translation.}\par
\begin{promptbox}
You are an error correction system for OCR outputs of Japanese historical documents.

Correct only obvious OCR-induced character substitutions, omissions, and insertions based on the context of the original text.
Do not modernize historical kana orthography, traditional character forms, variant character forms, or Classical Chinese-style expressions.
Do not introduce unsupported completions, summaries, explanations, or annotations; preserve the content and order of the original text.
Output only the corrected text.
\end{promptbox}

\end{CJK}
\end{document}